# Mentalic Net: Development of RAG-based Conversational AI and Evaluation Framework for Mental Health Support

Anandi Dutta
Texas State University
San Marcos, USA
anandi.dutta@txstate.edu

Shivani Mruthyunjaya
Texas State University
San Marcos, USA
aon9@txstate.edu

Jessica Saddington
*Texas State University*
San Marcos, USA
tgg55@txstate.edu

Kazi Sifatul Islam
Texas State University
San Marcos, USA
kazi_sifat@txstate.edu

*Abstract*— The emergence of large language models (LLMs) has unlocked boundless possibilities, along with significant challenges. In response, we developed a mental health support chatbot designed to augment professional healthcare, with a strong emphasis on safe and meaningful application. Our approach involved rigorous evaluation, covering accuracy, empathy, trustworthiness, privacy, and bias. We employed a retrieval-augmented generation (RAG) framework, integrated prompt engineering, and fine-tuned a pre-trained model on novel datasets. The resulting system, Mentalic Net Conversational AI, achieved a BERT Score of 0.898, with other evaluation metrics falling within satisfactory ranges. We advocate for a human-in-the-loop approach and a long-term, responsible strategy in developing such transformative technologies, recognizing both their potential to change lives and the risks they may pose if not carefully managed.

*Keywords—Large Language Model; retrieval-augmented generation (RAG); Conversational AI; Responsible AI*

## I. Introduction

Digital mental health solutions, particularly conversational Artificial Intelligence (AI), present a promising opportunity to address limitations in mental health support. In the USA, as of September 30, 2021, an estimated 126.9 million people lived in one of the professional shortage areas. There are about 5,930 federally designated areas identified as Mental Health Professional Shortage Areas[1]. Conversational AI, often implemented as chatbots, has the potential to bridge the growing gap between mental health needs and the capacity of available resources by providing accessible and convenient assistance not only in the USA but also worldwide. This technology can democratize access to mental health resources and can provide support in a stigma-free, 24-hour available environment. However, this open-ended technology must be developed with a long-term strategy and a human-in-the-loop approach to mitigate research challenges, such as those related to empathy, trustworthiness, privacy, and bias. In the first stage, we focused on developing the chatbot and performed rigorous evaluation, including technical testing, bias testing, and privacy/security testing, along with human evaluation. We proposed a workflow on how it can be integrated into the therapist's workflow, supported by data-driven analysis, ultimately augmenting the professional healthcare process. In the second phase, our strategy will extensively incorporate input and evaluations from behavioral health experts, therapists, and regular user feedback. This technology can potentially have a significant positive impact if managed and developed properly [24]. Eventually, the final goal is to develop a holistic evaluation framework for this technology that can ensure AI standards and quality assurance of this technology.

This is still a work-in-progress and proof-of concept project. The purpose of this project is to explore this open-ended domain, gain insights, and contribute to developing meaningful and safe technology in this emerging field. Moreover, our development targets the general public, aiming to support common well-being and mental health issues. This is not intended for handling severe issues or diagnosis.

## II. Literature Review

### A. Chatbot Development in the Healthcare Domain

The implementation of chatbots within varied contexts for different applications has shown very promising results in supporting individuals in achieving their corresponding aims. The "Habit Bot" is designed to support users in habit change through personalized interaction [20]. The chatbot enabled the clear definition of a chatbot's strong ability to support individuals in changing their unconscious routines or behaviors that don't serve them. Similarly, chatbots have been implemented within a medical context with the aim of providing logical support to medical professionals for better diagnosis and better patient care. Another study developed a personalized medical healthcare assistant, fine-tuned its model, applied prompt engineering, and utilized a robust RAG pipeline [17]. Another study evaluated mental health chatbots' sentiment response, but only through technical measures [15]. In another study, Schyff et al. [18] provided a roadmap for self-assessment and self-help utility utilizing Conversational AI. On the other hand, Loh et al. [19] demonstrated that large language models demonstrated promising results in generating empathetic responses in a comparative study. We conclude that further research is needed in this area due to its open-ended nature, the complexity of human behavior, and the nuances of language. Moreover, our study emphasizes both a thorough/robust technical evaluation and meaningful integration of human judgment in the development process. In this work, we primarily tested the technical robustness of our model with a limited human-in-the-loop approach. However, our next step is to extensively incorporate input and evaluations from behavioral health experts, therapists, and regular user feedback.

Recent years have seen the rapid adoption of retrieval-augmented generation (RAG) to make large language models (LLMs) more factual and context-aware in healthcare. In RAG, a retriever (e.g. BM25 or vector search) selects relevant documents or knowledge (from medical texts, guidelines, patient records, etc.), and an LLM conditions on that retrieved content to generate its response. By grounding generation in a trusted knowledge base, RAG can boost factual accuracy and reduce hallucinations [2]. This is especially important in sensitive domains like medicine and mental health, where incorrect or outdated advice can lead to serious consequences. Here, we focus on system architectures, retrieval mechanisms, fine-tuning strategies, datasets, evaluation metrics, and validation for the literature review.

Typical RAG pipelines combine a retriever and an LLM. For instance, Das et al.[3] designed a two-layer RAG system for medical question answering: the first layer retrieves relevant Reddit posts (via a simple BM25-based engine) and has the LLM summarize them; the second layer composes a final answer. Similarly, LLMind Chat [4] uses an open-domain retriever over the ICD-11 mental health taxonomy. Its RAG model is built on Gemma-2 (a 27B-parameter open LLM) and a custom ICD-11 knowledge base, so that the chatbot can fetch precise diagnostic information in real time. Silva et al. [5] implement MentalRAG, a multi-agent framework where one agent uses a RAG pipeline over clinical guidelines (stored in a ChromaDB vector store) to generate therapy proposals. In all cases, the retriever is a crucial component. Some systems (e.g. Das et al.) use keyword/BM25 search (e.g. Whoosh with Okapi BM25 ), while others employ vector embeddings (e.g. a ChromaDB store in MentalRAG). The choice of retrieval source reflects domain needs: LLMind Chat draws on the ICD-11 manual, Das et al. query Reddit mental-health forums, and MentalRAG retrieves from evidence-based psychiatric guidelines.

Most works start with a strong pre-trained LLM and fine-tune or prompt it for the healthcare domain. In LLMind Chat, a 27B-parameter Gemma-2 model is prompted (with retrieval) rather than heavily fine-tuned. Das et al. fine-tune or prompt large LLMs (GPT-4, Nous- Hermes, etc.) on clinical queries; their two-layer RAG uses off-the-shelf GPT models to summarize retrieved text, with no additional task-specific fine-tuning mentioned.

RAG systems retrieve from knowledge bases such as ICD-11, clinical guidelines, and user-generated posts. Datasets include DSM-5 cases, Reddit Q&A on psychoactive substances, and mental health surveys. These diverse sources enable the generation of a contextually rich and up-to-date information.

Metrics include BLEU, ROUGE, BERTScore, and human expert ratings. Evaluations measure factual accuracy, relevance, empathy, and safety. Human judgments are essential, especially for mental health applications.

In summary, RAG-based conversational agents in healthcare combine pre-trained LLMs with the retrieval of domain knowledge to improve accuracy and relevance. Evaluation combines automated metrics with expert review, showing potential for safe, empathetic mental health support. Challenges remain in ensuring data privacy, reducing biases, and maintaining real-world relevance.

Compared to other works, our work contributes to four key areas: Novel Dataset Creation; Development (combining RAG, fine-tuning, and prompt engineering); a holistic and rigorous Evaluation Framework; and a proposed Framework to augment the therapist's workflow.

III. METHODOLOGY

We developed a Retrieval-Augmented Generation (RAG) based Conversational AI system tailored for mental health support. First, we curated a comprehensive knowledge base comprising verified mental health resources, including clinical guidelines, self-help materials, and publicly available counseling datasets. We implemented a retriever component using a hybrid of BM25 and dense vector search (e.g., FAISS with sentence embeddings) to efficiently fetch relevant documents in response to user queries. For generations, we did fine-tune a pre-trained large language model (e.g., TinyLLaMA) using domain-specific dialogues and question-answer pairs, with retrieval outputs as context. We also used prompt engineering to enhance empathy and appropriateness. Evaluations included automated metrics (BLEU, ROUGE, BERT Score) and human evaluation to assess factual accuracy, relevance, and safety. We performed rigorous bias and privacy/security audits to ensure ethical deployment. Fig. 1 and Fig. 2 show the flowchart of the development process. We consider the "development" and "evaluation" process as adaptive, iterative, and concurrent.

A. Dataset Selection

Our primary focus was to ensure responses with emotionally empathetic and professional dialogue, in order to support users' sensitivity and conversational flow. We identified several datasets that not only contained high-quality human verbal interactions but also displayed a diversity of emotional situations, mental health relevance, and realistic conversational structure. The first dataset chosen was the Empathetic Dialogues Dataset from Facebook, AI, available at Kaggle. This dataset [6] includes 25,000 conversations grounded in specific emotional contexts, supporting a model's ability to handle a wide variety of emotional expressions or, at times, the appropriate level of apathy when needed. Most existing chatbots seem to be very affirming of user actions, however this dataset supports both the acknowledgement of user feelings but also allows for growth-oriented language as well.

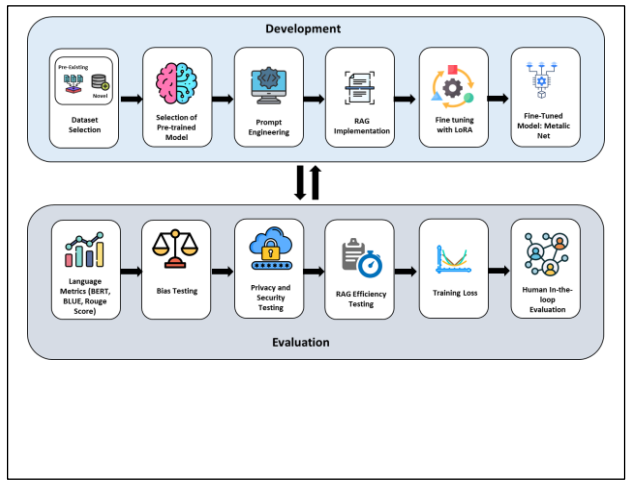

Fig. 1: A Flowchart Illustrating the Concurrent/Adaptive/Iterative Development and Evaluation Process

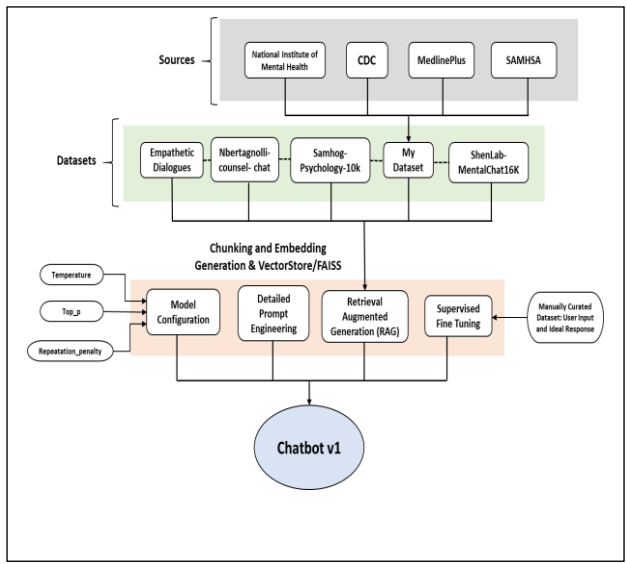

Fig. 2: Flowchart of Mentalic Net Conversational AI Development

The second dataset [7] was from GitHub, named "Counsel Thoughts". Due to its advice-oriented format, incorporation of context, and professional tone, it was deemed crucial to tune the chatbot. Most importantly, this dataset was anonymized from counselor responses to real-world therapeutic questions. The combination of the different datasets strongly supports the well-rounded fine-tuning of the selected model to ensure emotionally intelligent yet professional language.

The third dataset [8] implemented within the chatbot was created by us. There exist several different open-source, reputable websites that focus on providing information about mental health knowledge as well as advice. Several websites, such as the National Institute of Mental Health [8], CDC [9], MedlinePlus [10], and SAMHSA[11] were utilized to create this dataset.

The fourth dataset [12] was the "psychology-10k" which contained user questions and the fitting chatbot responses. The dataset consisted of about 10k dialogue examples crafted to support the development of conversational AI systems for mental health. The dataset presents a patient's input and an emotionally supportive psychiatrist's response. The fifth dataset that was chosen to be implemented was the "MentalChat16K[13]", which was found through Huggingface. This dataset consists of 6338 question-answer pairs derived from 378 anonymized interview transcripts collected during PISCES clinical trials. The transcripts documented interactions between behavioral health coaches and caregivers of patients in palliative or hospice care. The transcripts were paraphrased using the Mistral Model for the privacy of the individuals involved, prior to the release of the dataset. The rest of the available question-answer pairs were synthetically created utilizing OpenAI's GPT-3.5 Turbo model. The synthetic data covered a range of 33 different mental health topics, including anxiety, depression, relationships, and family conflict.

*B. Retrieval-Augmented Generation (RAG) Implementation*

To implement the Retrieval-Augmented Generation (RAG), all the selected datasets were formatted into a combined JSONL file, following a question and answer format. This was implemented by saving the file within a private google drive, and allowing the program to access the google drive through an external link to load in the required information during the chatbot setup.

There were several different reasons that supported our choice to utilize RAG within the context of a mental health conversation AI chatbot. The first was the known behavior and inclination of language models to "hallucinate" and respond with information that may not be professionally accurate or relevant for the given situation. But the application of RAG for the chatbot computation reduces the hallucinatory behavior of the language model. In addition, it is computationally conducive to allow the chatbot to retrieve the information that it finds most relevant for a specific situation, instead of training the model on all domain-specific data. It is immensely supportive that one does not need to entirely re-train a model to include more domain-relevant knowledge, and the knowledge can simply be updated.

We implemented Retrieval-Augmented Generation (RAG) to improve the accuracy, relevance, and scalability of our conversational AI. RAG comprises two key components: a retriever that extracts relevant information from an external knowledge base, and a language model that utilizes this information to generate responses. Due to the warm, professionally supported, and strong responses used for our knowledge base, we indirectly tuned the format and tone of the chatbot responses. This enables the system to deliver more accurate and contextually grounded answers while avoiding hallucinations often associated with generative models operating in isolation.

To prepare the knowledge base for the RAG retrieval pipeline, we implemented a document loading function that would process the large collection of conversation data stored in the json file. Each line in this file contains a question-answer pair.

The "answer" contains the technical, core content that would be retrieved during the inference to create the chatbot's response. The function reads each line, parses the JSON object, and creates a Document object using the answer as the main content (page_content) and the associated question as metadata under the "title" field. This structure ensures that the documents remain linked to their original context while prioritizing the factual content for indexing. To maintain the quality and the ease of implementation, the function filters out any documents that are decidedly short (less than 50 characters). This ensured that the remaining indexed answers in the knowledge base were meaningful and well-formatted text.

After filtering was completed, the documents were processed using the CharacterTextSplitter, which breaks longer answers into smaller, overlapping text chunks of 600 characters each, with about a 0 character overlap. The process of chunking supports the retrieval process by guiding the system to return more targeted segments of text in response to user questions, rather than entire document indexes. These split "documents" are then passed to the vector embedding pipeline for indexing and retrieval. To ensure that RAG operates with maximum effectiveness, undertaking data pre-processing and segmenting prior to embedding is vital.

The second aspect of the RAG implementation involves the response generation from the knowledge base available. When a user submits a query, the query is embedded into the same vector space, and the system retrieves the top-k, optimized to a value of 6, most relevant answers based on cosine similarity. These retrieved sections are then passed to a language model as contextual input, allowing it to generate a well-suited and contextually relevant response.

*C. Model Selection, Fine-Tuning Process, and User Data Encryption/Protection*

Several different models were considered to be utilized to be tuned to be a mental health chatbot; each had their own benefits and drawbacks. Some seemed to be too significantly large in size, and many were initially intended to be utilized to construct the chatbot, but it was identified that they could not be utilized for commercialization and distribution. Other attributes were also considered, such as those which were at least somewhat inclined towards coherent responses to regular questions. We decided to create the first instance of the chatbot towards eventual strong development, choosing a model that would have a strong foundation to build upon in terms of its conversational abilities, contextual awareness, and minimal size would be beneficial. As any future development after being perfected in its application could be replicable with larger or other models, possibly better suited.

In order to directly create a prototype of the chatbot, something tangible that could be reviewed and then further developed upon, we decided to implement the RAG protocol with the datasets above, as well as a prompted focus on the selected "TinyLlama-1.1B-Chat-v1.0" tuned version of the model. This was obtained from the specific Huggingface page of "TheBloke/TinyLlama-1.1B-Chat-v1.0-GGUF" [14].

The TinyLlama-1.1B-Chat-v1.0-GGUF model is a compact language model built upon the architecture and tokenizer of Llama 2. There were several reasons for the use of this model for chatbot development. This model was previously fine-tuned for conversational use, it is computationally lightweight, and it is fast. One key reason for this choice was the TinyLlama's transformer architecture, which is proven to be well-suited for modern language understanding and generation. Through its implementation of self-attention for sequence modeling, FlashAttention for efficiency on long inputs, and RoPE (Rotary Position Embeddings) for encoding sequence order, it is well suited to support the development of a chatbot.

We also deployed prompt engineering as part of our process. The prompt is mentioned below. It was considered to be during model output generation: "You are a compassionate, emotionally intelligent mental health assistant. Your role is to support users with warmth, honesty, and practical insight — like a trusted peer or counselor.
Instructions:
- Begin by validating the user's feelings in a natural, grounded tone. Avoid sounding robotic, clinical, or overly cheerful.
- Speak with empathy and clarity — like a calm friend who genuinely understands.
- Use casual, everyday language that feels safe and approachable.
- Keep responses concise (about 2–4 sentences) unless the user asks for more detail.
- When appropriate, offer **one clear, low-effort action** the user can try immediately (e.g., breathing technique, reflection prompt, or mental shift).
- Do NOT offer generic advice, toxic positivity, or clichés. Be specific, real, and kind.
- Never fabricate facts or cite sources not found in the given context.
- Only use provided context if it directly helps the user's current concern.
- Focus on soothing emotional distress, building trust, and offering useful next steps."

Within this format, we were able to receive appropriate responses for different user inputs. After having the RAG implementation effectively set up, we moved to identifying the best configuration parameters for the model. For the current state of the chatbot, the main aspects of fine-tuning that were undertaken were through RAG and the alteration of the configuration parameters. The process was undertaken to ensure that the model was providing contextually relevant responses and was providing professionally supported mental health advice to users.

The process was undertaken by initially setting each parameter to a safe estimated value, and through iterative use of the chatbot, we were able to identify and optimize each parameter. Initially the configuration parameters were a max_new_tokens value of 600, a temperature of 0.7, a top_p of 0.9, a repetition_penalty of 1.3, and a context_length" of 2048. In order to minimize the length of the chatbot responses for conciseness, the max_new_tokens was lowered to 500. The temperature parameter which affects the "randomness" of the output chosen, was lowered so that there was more predictability. The "top_p" parameter was significantly lowered such that the conversational creativity more closely followed the provided conversation format. This allowed for more focused and less diverse outputs. The repetition penalty was lower, allowing the model to elaborate further on prior statements without restriction. The final model configuration was chosen to be a "max_new_tokens" value of 500, a "temperature" value of 0.65, a "top_p" of 0.3, a repetition_penalty of 1.1, and finally the same context_length of 2048.

These choices allowed for a well performing mental health chatbot with emotionally supportive language and professionally approved mental health resources/advice.

Finally, we implemented an external secure storage process of the encrypted version of the user's chat history and output, for a specific encrypted input. This process was completed using a webhook_url for a Google Sheet, and functions were created to encrypt data properly. We ensured that we captured the time and date of the interaction with the chatbot. This information was stored within a Google Sheets document.

IV. EVALUATION OF THE DEVELOPED CONVERSATIONAL AI

Following the completion of the first version of the chatbot, a rigorous evaluation process was undertaken to ensure its reliability, effectiveness, and privacy assurance. We generated/created a 500 samples dataset for the evaluation process. The dataset contains User Question, Chatbot's Response, and Standard/Ideal Response. All the standard/ideal responses are created by humans with meticulous care and supported by medical texts. The language quality metrics of the chatbot were assessed by the BERT Score. BERT Score captures (Fig. 3) meaning rather than exact words, which is crucial for mental health applications where empathy and contextual appropriateness matter. The score is 0.898, which indicates that the chatbot's outputs are highly semantically similar to the ground-truth references.

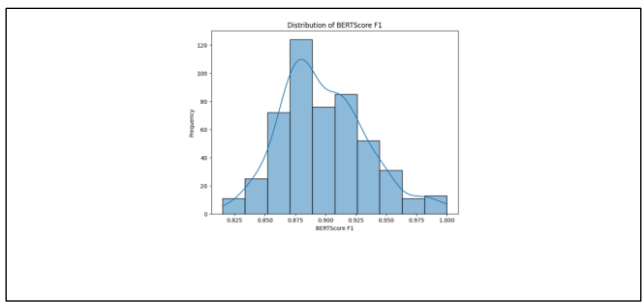

Fig. 3: Distribution of BERT score

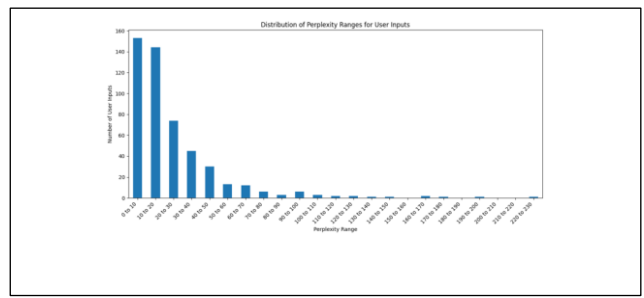

Fig. 4: Distribution of Perplexity score

To evaluate the effectiveness of the RAG retrieval, specific questions from the RAG were chosen, then, utilizing the chosen method, the "relevant" documents were retrieved. The precision of retrieving the "correct" documents and other relevant documents was recorded. The Precision and Recall scores of RAG Efficiency Testing are 0.51 and 0.86. A recall score of 0.86 demonstrates that the retriever successfully found 86% of all the truly relevant documents in the knowledge base. This high recall ensures that most of the necessary information is readily available for the language model, thereby reducing the likelihood of omitting key content in responses. We prioritized achieving a high recall score to ensure no critical information is missed. However, we aim to improve the precision score to at least 0.7.

While a precision score of 0.51 may initially suggest limited retrieval accuracy, the interpretation of the score must consider the context of the open-ended and conversational nature of the knowledge base implemented. A manual observation and thorough review of the retrieved documents demonstrated that they contained contextually relevant information aligned with the user queries, despite not being exact matches to all entries in the ground truth. The visual review of the retrieved documents showed that they provided useful content embedded within natural language structure, reflecting situational significance, tonal appropriateness, or responses partially aligned with the semantic meaning of the documents within the selection of ground truth documents. Therefore, we are able to confirm the practical effectiveness of the retrieval system, despite the precision score underrepresenting its true performance.

Our conversational AI system achieved a BERT score of 0.898 reflecting a high level of semantic accuracy, surpassing the performance of all the categorized BERTScore F1 presented in [23], derived from testing another constructed mental health chatbot. In comparison to generative models, our model outperforms DialogGPT (F1 = 0.7603) and GPT-2 (F1=0.7445) as discussed in [22], indicating comparatively strong contextual semantic accuracy. Another study focused on constructing a conversational AI system completed testing using several benchmarks, including the MELD, MultiDialog, and IEMOCAP [21]. Upon comparison, we can see that our model achieved a greater score BERT during evaluation with our constructed "Standard Ideal/Response," demonstrating the strength of our model in achieving semantic relevance during response generation.

We performed the Perplexity testing. Perplexity scores (Fig. 4) measure the model's ability to predict a sequence of words, with lower perplexity indicating better predictive performance. Perplexity is the exponential of the average negative log-likelihood of the correct token. We conducted this test on 500 samples, and it provided a perplexity score for each "user input". It provided us with better insights into which "user inputs" surprised our model. The average perplexity score is 25. However, we identified some questions such as "What is my love language?" "Can people truly change their core personality?" "How are you today?" "How do early attachment styles affect my relationships?" "My physical pain makes it hard to stay positive." demonstrated high perplexity scores (100 to 200). However, for most of the user inputs, the perplexity score is low (0 to 20).

| Evaluation Metrics | Scores/Satisfactory |
|---|---|
| BERT Score | 0.898 |
| Perplexity Score | ``` 0 to 10      153
10 to 20     144
20 to 30      74
30 to 40      45
40 to 50      30
50 to 60      13
60 to 70      12 ``` Here, the first column is the perplexity score and the second column is the number of user inputs/questions |
| RAG Efficiency Score | Recall: 0.86<br>Precision: 0.51 |
| Fine-Tuning Loss | 3.68 (training); 3.40 (validation); *This can be improved with larger dataset |
| Response Time | For 50 questions, the average response time was 13.8 |
| Load Testing Simulation | Completed 50 requests in 6.49 seconds |
| PII Leakage Detection | Satisfactory |
| Prompt Injection Testing | Satisfactory |
| Bias Testing with Human Evaluation | Satisfactory |

Table 1: Evaluation Metrics of Mentalic Net

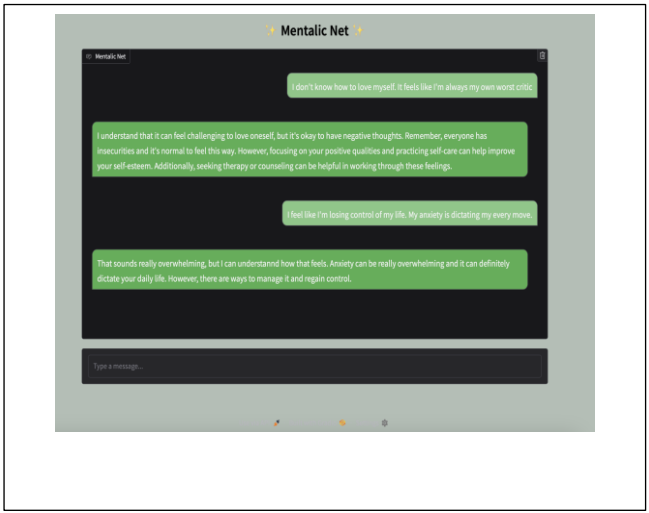

Fig. 5: Mentalic Net Interface: A Conversationl AI for mental health support and integration/augmentation to Therapist's Workflow

In addition to the other evaluations, bias and fairness checks were conducted to identify and mitigate any issues with chatbot's responses stemming from any cultural/religious/gender/demographic context. Human evaluation was performed for bias and fairness testing with specific curated questions and answers. It turned out to be satisfactory.

The security and privacy of the chatbot were evaluated through implementing PII leakage detection, prompt injection, and load testing simulations to assess system behavior with several users and increased usage length. All the tests turned out satisfactory. The various evaluations provided insights into potential improvements for the chatbot's next iteration and an assessment of its current quality. Table 1. Demonstrates detailed compilation of all the evaluation metrics.

## V. INTEGRATION INTO THERAPIST'S WORKFLOW

We would also like to propose the possibility of the integration of this chatbot into the therapist's workflow. This chatbot can be deployed as a smartphone/web application. The current version is a web version (Fig. 5). This provides the possibility of tracking/monitoring the user's conversation log and providing meaningful data-driven insights to the therapist. Due to the scarce resources and long appointment/waiting time, the digital measure of remote monitoring could be a very helpful alternative. Eventually, it can also alert the professionals/caregivers if a person's mental health deteriorates and encourages the user to seek immediate professional support. We created a synthetic dataset to demonstrate this approach. In Fig. 6, the calendar plot can be created based on the sentiment analysis of conversation logs and shared with the therapist based on the user's permission. In Fig. 7, a radar chart was created based on three months' conversation logs. The radar chart presents a comparative analysis of average sentiment scores for various emotions across January, February, and March. The emotions are mapped along the axes, including happiness, hopefulness, motivation, neutrality, sadness, tiredness, anger, and anxiety. The chart reveals consistently high sentiment scores for "happy" and "hopeful" emotions across all months, suggesting a generally positive mood trend. Conversely, "angry" and "sad" sentiments display consistently low scores, indicating these emotions were less prominent. Interestingly, February shows a slight increase in motivation compared to January and March, while March exhibits slightly higher levels of sadness and tiredness. This visualization highlights both the stability and subtle shifts in emotional sentiment over the three-month period, offering insights into patterns that may reflect seasonal influences or changes in external factors affecting emotional well-being. This is a demonstration of how this Mentalic Net Conversation AI has the possibility of augmentation of the therapist's workflow, if not a better understanding of self-assessment and state of mind. It can track a user's mood over a long period of time, until the person finally has the opportunity/will to receive professional help.

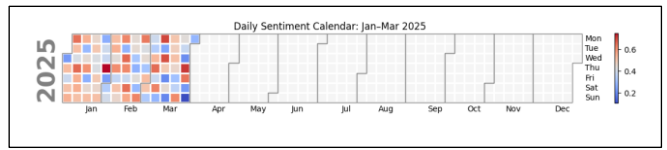

Fig. 6: Remote monitoring of User's Mental Health based on Conversation Logs' Sentiment Analysis

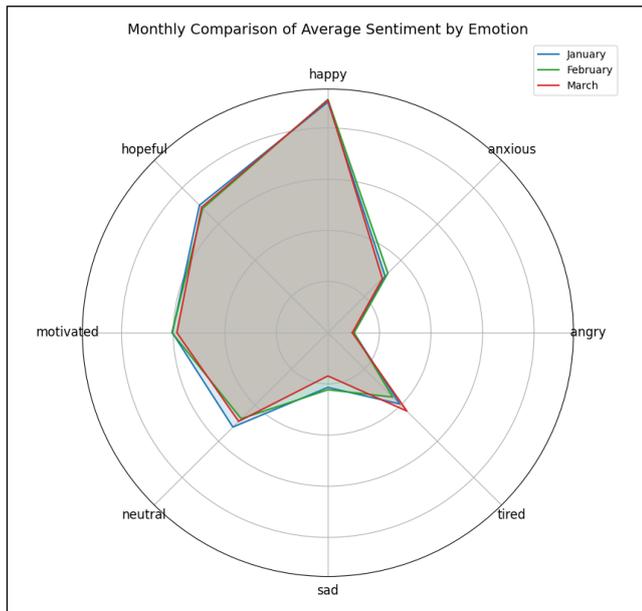

Fig. 7: Remote monitoring of User's Mental Health based on Conversation Logs' Sentiment Analysis

## VI. Conclusion

Again, this is the first version of the chatbot. The initial results demonstrated the strong potential of an effective tool for both general users and healthcare professionals. The reported BERT score is 0.898, and the reported RAG score is 0.51. Moreover, all initial load testing, security testing, and response time came out as satisfactory. The perplexity scores are acceptable. We performed human evaluation for bias testing. For the first version, we considered 500 samples, but we will increase the sample size for later versions.

In the next phase, we will perform a detailed evaluation with a selected general user focus group. Moreover, we will include behavioral health researchers and professional therapists' evaluations. Based on the feedback from these crucial evaluations, we will update and improve the currently developed conversational AI. Moreover, we are interested in extending our research on the disadvantages of this kind of technology, such as digital fatigue, alienation from friends and family, and over-dependence. This work is part of a long line of future work and is still in the developmental phase.